\documentclass{article}

\usepackage{microtype}
\usepackage{graphicx}
\usepackage{subfigure}
\usepackage{booktabs} 

\usepackage{hyperref}

\usepackage[accepted]{icml2019}
\usepackage{lipsum}
\usepackage{helvet}
\usepackage{courier}
\usepackage{bbm}
\usepackage{color}
\usepackage{makeidx}  
\usepackage{multirow}      
\usepackage{epstopdf}
\usepackage{amsmath,dsfont}
\usepackage{amsfonts}
\usepackage{epstopdf}
\usepackage{enumitem}
\usepackage{color}
\usepackage{url}
\usepackage{algorithm}
\usepackage{bbding}
\usepackage{wasysym}
\usepackage{colonequals}
\usepackage{mathrsfs}
\usepackage{lscape}
\usepackage{setspace}
\usepackage{pifont,textcomp,amssymb}
\usepackage{algorithm,algorithmic}
\usepackage{amsmath,calc}
\usepackage{multicol}
\usepackage{booktabs}

\newtheorem{theorem}{Theorem}

\newtheorem{definition}{Definition}

\newcommand{\nonl}{\renewcommand{\nl}{\let\nl\oldnl}}
\newcommand{\fin}{f_\iota}
\newcommand{\Di}{\mathcal{D}_\iota}
\newcommand{\Db}{\mathcal{D}_b}
\newcommand{\fb}{f_b}
\newcommand{\mR}{\mathcal{R}}

\icmltitlerunning{Gaining Free or Low-Cost Transparency with Interpretable Partial Substitute}

\begin{document}

\twocolumn[
\icmltitle{Gaining Free or Low-Cost Transparency with Interpretable Partial Substitute}


\icmlsetsymbol{equal}{}

\begin{icmlauthorlist}
\icmlauthor{Tong Wang}{ed}
\end{icmlauthorlist}

\icmlaffiliation{ed}{Department of Business Analytics, University of Iowa, Iowa, USA}

\icmlcorrespondingauthor{Tong Wang}{tong-wang@uiowa.edu}

\icmlkeywords{Machine Learning, ICML}

\vskip 0.3in
]



\printAffiliationsAndNotice{} 

\begin{abstract}
This work addresses the situation where a black-box model with good predictive performance is chosen over its interpretable competitors, and we show interpretability is still achievable in this case. Our solution is to find an interpretable substitute on a subset of data where the black-box model is \emph{overkill} or nearly overkill while leaving the rest to the black-box.  This transparency is obtained at minimal cost or no cost of the predictive performance. Under this framework, we develop a Hybrid Rule Sets (HyRS)  model that uses decision rules to capture the subspace of data where the rules are as accurate or almost as accurate as the black-box provided. To train a HyRS, we devise an efficient search algorithm that iteratively finds the optimal model and exploits theoretically grounded strategies to reduce computation. Our framework is \emph{agnostic} to the black-box during training. Experiments on structured and text data show that HyRS obtains an effective trade-off between transparency and interpretability.
\end{abstract}

\section{Introduction}
The deployment of machine learning in real-world applications has led to a surge of interest in systems optimized not only for task performance but also model interpretability, especially when human experts are involved in the decision-making process. In many heavily regulated industries such as judiciaries and health care, understanding the decision-making process of an analytical model is not just a preference but often a matter of legal and ethic compliance \cite{doshi2017roadmap}, as recently promoted by EU's General Data Protection Regulation (GDPR) \cite{parliament}.

Thanks to this increasing interest, interpretable machine learning has achieved unprecedented advancement, providing two solutions for facilitating human understandability.

The first solution is to develop models that are interpretable by themselves, such as rule-based models, scoring models, case-based models, etc., of small or reasonable sizes. They are self-contained and do not rely on other models to explain them. We call them \emph{interpretable} models in this paper and call the decision-making process \textbf{transparent} since humans understand wholly and precisely how a decision is generated. This solution is favorable when interpretable models perform as well as or better than black-box models \cite{choi2016retain}. However, due to possible constraints on the model complexity to achieve interpretability, the lose of predictive performance for choosing interpretable models is often inevitable \cite{wang2015trading},  needing another solution to obtain interpretability in this case. 

To deal with situations when an interpretable model is inadequate, and a black-box has to be chosen for better predictive performance, the second solution is proposed to develop models that explain black-boxes. These models generate post hoc explanations or approximations for a black-box either locally \cite{ribeiro2016should} or globally \cite{adler2016auditing,lakkaraju2017interpretable}, providing some insights into the black-box model by identifying key features or interactions of features \cite{tsang2017detecting}.
However, two concerning issues exist. First, explainers only approximate but do not characterize exactly the decision-making process of a black-box model, yielding an imperfect explanation fidelity. 
Second, there exists ambiguity and inconsistency \cite{ross2017right,lissack2016dealing} in the explanation since there could be different explanations for the same prediction generated by different explainers, or by the same explainer with different parameters. Both issues result from the fact that the explainers only approximate in a post hoc way but are not the decision-making process themselves.
 
In this paper, we create an alternative solution to gain interpretability in the presence of a chosen black-box.  We propose to use interpretable partial substitute to process a subset of data, where an interpretable model is adequate for producing predictions that are as good as the black box, i.e., where the black-box is overkill, to obtain free interpretability at no cost of the predictive performance; or, if the user is willing to trade some accuracy for transparency, our model can find the right subset of data at minimal cost of predictive performance. Thus, on this subset of data, the model gains transparency with 100\% fidelity to replace otherwise non-perfect approximations by an explainer. We define the percentage of the subset \emph{transparency} of the model.

To summarize, we design a novel framework that integrates an interpretable model with the black-box model into a sequential decision-making process.  An input instance first goes through an interpretable model. If the model is competent, a prediction will directly be generated. Otherwise, the black box will be activated.  We call the proposed model a \emph{Hybrid Predictive Model}. Under this framework, we build Hybrid Rule Sets (HyRS) where we use association rules as interpretable local substitutes. 

This form of the model is motivated by how humans make decisions in many real-world situations. For example, when a doctor diagnoses a patient, if the patient is an irregular case with symptoms that do not match documented descriptions of any disease, then an experienced doctor or a consultation of several experts will be summoned, representing a complicated model (black-box model). However, if the patient demonstrates regular ``textbook'' symptoms, a diagnosis can be made right away via standard symptom matching and reasoning (interpretable model) by a simpler ``model'' such as a resident or a nurse.  In these cases, residents and nurses can perform nearly well as experienced doctors since there are symptoms easy to explain.  For a hospital, it is not economical to always request a consultation from several experts to treat simple cases. Thus, hospitals often stratify patients and send more complicated cases to more experienced doctors. Our model works similarly, and our goal is to construct an interpretable model that replaces the complicated black-box model on an appropriate subspace of data. 

The benefits of a hybrid model include several aspects. \emph{First}, it gains transparency of the model at no cost or minimal cost of predictive performance. Compared to explainers that provide post hoc analysis, an interpretable substitute is understandable by itself. There does not exist any ambiguity or inconsistency since an interpretable model is entirely faithful to itself.  \emph{Second}, rules only use a small set of features while the black-box model needs to use all. In some applications where features are costly to get (e.g., medical test results in hospitals), using a small set of features reduces cost.  \emph{Finally}, from an operational viewpoint, a black-box model can be a cumbersome system with high costs in operating and maintaining. Using a simpler interpretable model will save computing resources and time in real applications.

In this paper, we formulate a general framework and learning objective for building a hybrid model. The objective considers predictive accuracy, transparency, and interpretability.  The model learns to capture the correct subset of data and achieves the best balance between sending enough data to the interpretable model and preserving predictive performance. To train the model, we devise an efficient search algorithm that exploits the theoretical properties of the model to reduce computation complexity. We develop three bounds to reduce the search space, one applied before the search begins and two dynamic bounds during the search. 

In the rest of the paper, we will review related work in Section \ref{sec:related}. We present the general framework for learning a hybrid decision model in Section \ref{sec:framework} and develop a Hybrid Rule Sets (HyRS) model under the proposed framework in Section \ref{sec:model}. We design an efficient training algorithm that exploits theoretically grounded strategies for fast computation in Section \ref{sec:search}. Finally, we provide a detailed experimental evaluation in Section \ref{sec:exp}. We  conclude the paper in Section \ref{sec:conclusion}.

\vspace{-2mm}
\section{Related Work}\label{sec:related}
Our work is broadly related to new methods for interpretable machine learning.  
 There have been two lines of research in interpretable machine learning. The first is developing models that are interpretable stand-alone.  Previous work in this category include rule-based models such as fuzzy rules \cite{alonso2018bibliometric},  rule sets \cite{rijnbeek2010finding,mccormick2011hierarchical} and rule lists \cite{ynormalize_addang2016scalable,angelino2017learning}), scoring systems \cite{zeng2017interpretable,ustun2016supersparse,koh2015two}, and etc. 
The second line of research is on developing explainer models that explain black-boxes locally \cite{ribeiro2016should} or globally \cite{adler2016auditing,lakkaraju2017interpretable}. One representative work is LIME \cite{ribeiro2016should} that explains the predictions of any classifier by learning a linear model locally around the prediction.  More recently, developments in deep learning have been connected strongly with interpretable machine learning and have
contributed novel insights into representational issues. Recent works have proposed high-level symbolic representations used in knowledge representation \cite{yi2018neural}.

Our work is fundamentally different from the research above. A hybrid decision model is not a pure interpretable model. It uses an interpretable substitute on a subset of data. It is also not a diagnostic model that only observes but does not participate in the decision process. Here, a hybrid decision model uses an interpretable and a black-box model simultaneously in decision making and utilizing the strength of interpretable models in producing understandable predictions.  

There exist a few singleton works that combine multiple models. For example, \cite{kohavi1996scaling} proposed NBTree which induces a hybrid of decision-tree classifiers and Naive-Bayes classifiers, \cite{shin2000hybrid} proposed a system combining neural network and memory-based learning, \cite{hua2006hybrid} combined SVM and logistic regression to forecast intermittent demand of spare parts, etc. A recent work \cite{hybrid2018} builds a black-box oracle which outputs the simplest interpretable model to produce a similar prediction a black-box model would generate. There exist potential inconsistency issues since there are multiple interpretable models likely to be selected. Another work \cite{wang2015trading} divides feature spaces into regions with sparse oblique tree splitting and assign local sparse additive experts to the individual areas.
Our model is distinct in that the proposed framework can work with \textbf{\emph{any}} black-box model and is \textbf{\emph{agnostic}} to the model during training. 

 \vspace{-2mm}
\section{A General Framework for Hybrid Models}\label{sec:framework}
We present a general framework for building a hybrid model and define a principled objective function. 
We start with a set of training examples $\mathcal{D} = \{(\mathbf{x}_i,y_i)\}_{i=1}^N$ where $\mathbf{x}_i\in\mathcal{X}$ is a tuple of attributes and $y_i\in\{1,0\}$  is the class label. 
Let $f=\langle    \fin,\fb\rangle    $ represent a hybrid model that consists of an interpretable model $f_\iota$ and a black-box model $f_b$. $f$ is agnostic to $f_b$ and only needs the predictions of $f_b$  on $\mathcal{D}$ as input, denoted as $\mathcal{Y}_b =\{\hat{y_b}_i\}_i^N$.

\begin{figure}[h!]
\centering
  \includegraphics[width=0.46\textwidth]{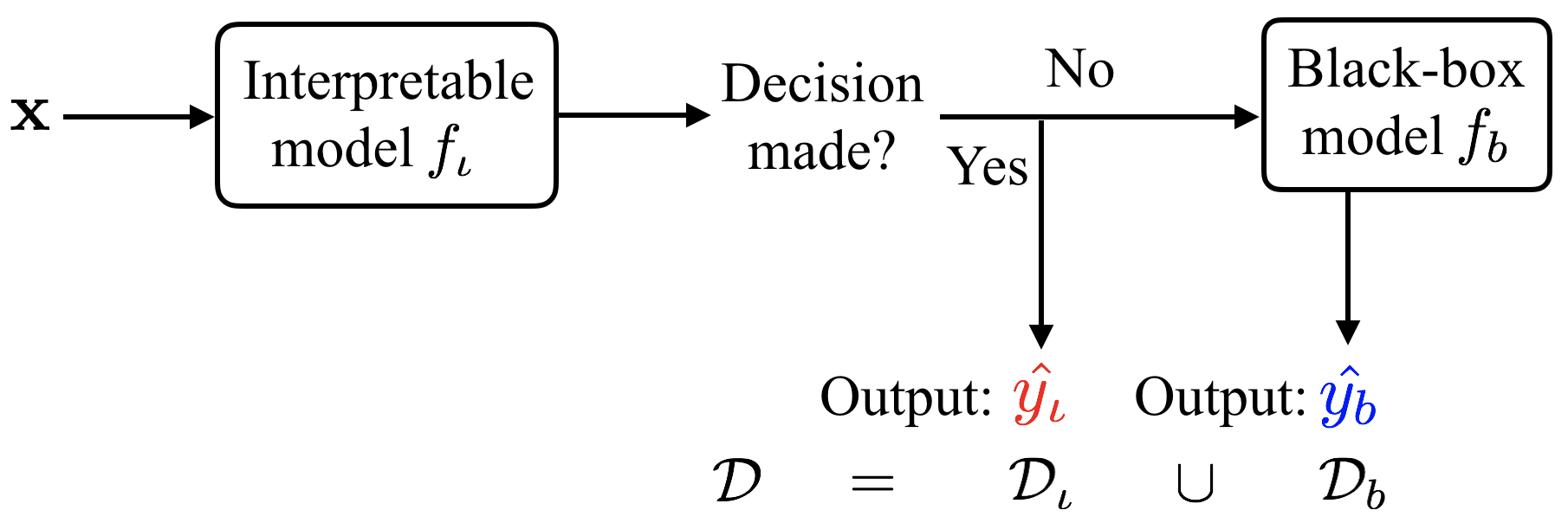}
\caption{A predictive process of a hybrid model.}\label{fig:hybrid}    
\end{figure}

 A critical issue in designing a hybrid model is how to automatically distribute data to $\fin$ and $\fb$. This is equivalent to creating a partition of the dataset $\mathcal{D}$ to $\Di$ and $\Db$, corresponding to training examples sent to $\fin$ and $\fb$, respectively. We design the predictive process as below: an input instance $\mathbf{x}_k$ is first sent to the interpretable model $\fin$. If a prediction can be made, an output $\hat{y_\iota}_k$ is directly generated. Otherwise, it is sent to $\fb$ to generate a prediction $\hat{y_b}_k$. $\hat{y_\iota}_k \in \Di$ and $\hat{y_b}_k \in \Db$. See the predictive process in Figure~\ref{fig:hybrid}.

Our goal is to construct an interpretable model $f_\iota$ to be combined with $f_b$, with the objective that considers three fundamental properties: 1) The \textbf{predictive accuracy}. Since $\fb$ is pre-given, the predictive performance of $f$ depends on the predictive accuracy of $\fin$ independently and the collaboration of $\fin$ and $\fb$, i.e., the partition of $\mathcal{D}$ to $\Di$ and $\Db$. $\fin$ and $\fb$ being completely different models allows the hybrid model to exploit the strengths of both models if the training examples are partitioned strategically, sending examples to the model which can predict them correctly. In some circumstance, the combination of a weak and a strong model can yield performance better than the strong model alone. 2) The \textbf{interpretability} of $\fin$. Bringing interpretability into the decision process is one of the motivations for building a hybrid model. Therefore, small size and low complexity are much-desired properties of $\fin$. The definition of interpretability is model specific and usually refers to using a small number of cognitive chunks \cite{doshi2017roadmap}.  
3) The \textbf{transparency} of the hybrid model. This is a new metric we propose for the hybrid framework to capture the percentage of data that are processed by $\fin$, i.e., the percentage of $\Di$ in $\mathcal{D}$. 
\vspace{-2mm}\begin{definition}\label{def:exp}
The transparency of a hybrid model $f = \langle \fin,\fb \rangle$ on $\mathcal{D}$ is the percentage of data processed by $\fin$, i.e., $\frac{\Di}{\mathcal{D}}$, denoted as $\mathcal{T}(f,\mathcal{D})$.    
\end{definition}\vspace{-2mm}
We formulate the learning objective for building a hybrid decision model as a linear combination of the three metrics above. 
This framework unifies interpretable models and black-box models: interpretable models have transparency of one, and black-box models have transparency of zero.

\vspace{-2mm}
\section{Hybrid Rule Sets Model}\label{sec:model}
Under the proposed framework, we instantiate a hybrid decision model. Here we take a significant step towards interpretability by choosing rules for $f_\iota$. Rules are easy to understand due to their simple logic and symbolic presentation. They also naturally handle the partition of data by separating examples according to if they satisfy the rules.

Now we present the Hybrid Rule Sets (HyRS) model. A HyRS model consists of two sets of rules. The first set of rules captures positive instances, called the \emph{positive rule set} and denoted as $\mathcal{R}_+$. The second set of rules captures negative instances, called the \emph{negative rule set} and denoted as $\mathcal{R}_-$.  If $\mathbf{x}_k$  satisfies any positive rules, it is classified as positive. Otherwise, if it satisfies any negative rules, it is classified as negative. Denote $\mathcal{R} = \mathcal{R}_+ \cup\mathcal{R}_-$. A decision produced from $\mathcal{R}$ is denoted as $\hat{y_\iota}_k$. If $\mathbf{x}_k$ does not satisfy any rules in $\mathcal{R}_+$ or $\mathcal{R}_-$, it means $\fin$ fails to decide on $\mathbf{x}_k$. Then $\mathbf{x}_k$ is sent to the black-box model $f_b$ to generate a decision $f_b(\mathbf{x}_i)$. $\Di$ is a set of instances sent to $\fin$. In the context of a HyRS model, we use $\fin$ and $\mR$ interchangeably when we refer to the interpretable model. We summarize the decision-making process below.
\begin{align*}
&\textbf{if } \mathbf{x}_i \text{ obeys } \mR_+, Y = 1 \notag \\
&\textbf{else if }\mathbf{x}_i \text{ obeys } \mR_-, Y = 0 \notag \\
&\textbf{else }Y= f_b(\mathbf{x}_i)
\end{align*}
We show an example of a HyRS model in Table~\ref{exp1} learned from a heart disease dataset from UCI ML repository \cite{Lichman:2013}. In this model, there are two rules in $\mathcal{R}_+$ and one rule in $\mathcal{R}_-$. $Y=1$ represents the patient has heart disease.
\vspace{-2mm}\begin{table}[h!]
\centering
\small
\caption{An example of a HyRS model}
\label{exp1}
\begin{tabular}{lll}
\toprule
        & \multicolumn{1}{c}{\textbf{Rules}}                                                             & \multicolumn{1}{c}{\textbf{Model}}             \\\hline
\textbf{if}      & age$<$ 35 \emph{and} maximum heart rate $\geq 178 $                               & \multicolumn{1}{c}{$\mathcal{R}_+$} \\
        & OR serum cholestorol $\geq 234$ \emph{and} thal $\neq 3$                           &                                       \\
        & \emph{and} the number of vessels $\geq 1$                                                        &                                       \\
        & $\rightarrow$ \textcolor{red}{$Y = 1$} (heart disease)             &                                       \\
\textbf{else if} & chest pain type $\neq 4$ \emph{and} age $>40$                                                &  \multicolumn{1}{c}{$\mathcal{R}_-$}              \\
        & $\rightarrow$ \textcolor{red}{$Y = 0$} ( no heart disease) &                                       \\
\textbf{else}    & $\rightarrow$ \textcolor{blue}{$Y = \fb(\mathbf{x})$}                             &  \multicolumn{1}{c}{$f_b$}      \\ \bottomrule                
\end{tabular}\vspace{-2mm}
\end{table}

To formulate the problem, we need the following notations and definitions.
\begin{definition}
A rule $r$ \emph{covers} an example $\mathbf{x}_i$ if $\mathbf{x}_i$ obeys the rule, denoted as $\text{covers}(r,\mathbf{x}_i) = 1$.
A rule set $R$ \emph{covers} an example $\mathbf{x}_i$ if  $\mathbf{x}_i$ obeys at least one rule in $R$, i.e. 
$
\text{covers}(R,\mathbf{x}_i) = \mathbbm{1}\left(\sum_{r\in R}\text{covers}(r,\mathbf{x}_i)\geq 1\right).
$
\end{definition}
\begin{definition}\label{def:supp}
Given a data set $\mathcal{D}$, the \textit{support} of a rule set $R$ in $\mathcal{D}$ is a set of observations covered by $R$, i.e., $\text{support}(\mathcal{R},D) = \{i|\text{covers}(\mathcal{R},\mathbf{x}_i) = 1,\mathbf{x}_i\in \mathcal{D}\}$.
\end{definition}

We formulate the objective function for HyRS following the objective described in the previous section. First, we measure the misclassification error to represent the predictive performance. Given rules $\mathcal{R}$, a black-box model $\fb$ and data $\mathcal{D}$, the misclassification error is

\small\vspace{-4mm}
\begin{align*}
&\ell(\langle\mathcal{R},\fb\rangle,\mathcal{D})= \sum_{i=1}^N \bigg((1-y_i)\text{covers}(\mathcal{R}_+, \mathbf{x}_i)\!\!\! &&\!\!\triangleright\text{errors from }\mathcal{R}_+ \notag \\
&+y_i\Big(1-\text{covers}(\mathcal{R}_+, \mathbf{x}_i)\Big)\text{covers}(\mathcal{R}_-,\mathbf{x}_i)\!\!\!  &&\!\!\triangleright      \text{errors from }\mathcal{R}_- \notag \\
&+(1-\text{covers}(\mathcal{R}_+, \mathbf{x}_i)(1-\text{covers}(\mathcal{R}_-, \mathbf{x}_i) \!\!\!&&  \notag \\
&\times\big(y_i(1-\hat{y_b}_i) + (1-y_i)\hat{y_b}_i\big) \bigg)/N.\!\!\! &&\!\!\triangleright\text{errors from }\fb  \label{eqn:error}
\end{align*}
\normalsize
\vspace{-4mm}
\begin{definition}\label{def:size}
$\Omega(R)$ is the interpretability of $R$.    
\end{definition}
There are various definitions of interpretability for rule-based models, such as the number of rules \cite{NIPS2018_7716,wang2017bayesian,lakkaraju2016interpretable,NIPS2018_8281}, the number of conditions, and those discussed in \cite{gacto2011interpretability}. The definition of interpretability is domain specific and subject to applications. In this paper, we choose the number of rules in $R$ to represent interpretability as an illustration of the framework but our formulation works with other interpretability measures in the literature.

We aim to minimize the objective function combining predictive accuracy, model interpretability and transparency.
\begin{equation}\label{eqn:obj}
\scriptsize
\Lambda(\langle\mathcal{R},\fb\rangle,\mathcal{D}) = \ell(\langle\mathcal{R},\fb\rangle,\mathcal{D}) + \theta_1\Omega(\mathcal{R}) -\theta_2 \frac{\text{support}(\mathcal{R},\mathcal{D})}{N}
\end{equation}\normalsize
where the transparency of a HyRS model follows definition~\ref{def:exp}.
Here, $\theta_1$ and $\theta_2$ are non-negative coefficients. Tuning the parameters will produce models at different operating points of accuracy, interpretability, and transparency. For example, in an extreme case when $\theta_2 >> \theta_1$, the output will be a model that sends all data to $\fin$, producing a pure interpretable model. When $\theta_1>>\theta_2$ and $\theta_1 >>1$, then the model will force $\fin$ to have complexity 0, i.e., producing a pure black-box model.

\vspace{-2mm}
\section{Model Training}\label{sec:search}
We describe a training algorithm to find an optimal solution $f^*$ such that 
\begin{equation}
    f^* \in \arg\min_{f} \Lambda(f; \mathcal{D})
\end{equation}
Since $f^* = \langle \mR^*, f_b \rangle$ and $f_b$ is fixed, the problem reduces to finding an optimal rule set model $\mR^* = \mR^*_- \cup \mR^*_-$ that covers the correct subset of data in the presence of $\fb$.

 Learning rule-based models is challenging because the solution space (all possible rule sets) is a power set of the rule space. Fortunately, our objective has a nice structure that can be exploited for reducing computation.

\vspace{-2mm}
\paragraph{Algorithm structure} The algorithm is presented in Algorithm 1. Given training examples, $\mathcal{D}$, a black-box model $\fb$, parameters $\theta_1,\theta_2$,  base temperature $C_0$ and the total number of iterations $T$, the search procedure follows the main structure of a stochastic local search algorithm. Each state corresponds to a rule set model, indexed by the time stamp $t$, denoted as $\mR_{[t]}$. The temperature is a function of time $t$, $C_0^{1-\frac{t}{T}}$, and it decreases with time.  The neighboring states are defined as rule sets that are obtained via adding or removing a rule from the current set. At each iteration, the algorithm improves one of the three terms (accuracy, interpretability, and transparency) with approximately equal probabilities, by removing or adding a rule to the current model.
The proposed neighbor is accepted with probability $\exp(\frac{\Lambda( \mathcal{R}_{[t]}) - \Lambda( \mathcal{R}_{[t+1]}) }{C_0^{1-\frac{t}{T}}})$ which gradually decreases as the temperature cools down. 
\begin{algorithm}[h!]\label{alg:search}
\caption{Stochastic Local Search algorithm}
\small
\begin{algorithmic}[1]
\STATE \textbf{Input:} $\fb, \mathcal{D}, \theta_1,\theta_2, C_0$
\STATE \textbf{Initialize:}  
            \STATE $\mathcal{R}^{*}  = \mathcal{R}^{[0]} \leftarrow \emptyset$ $ \triangleright$ start with a pure black-box model
           \STATE $\Upsilon^+ \leftarrow \text{FPGrowth}(\mathcal{D}, \text{minsupp} = N\theta_1)$
            \STATE $\Upsilon^- \leftarrow \text{FPGrowth}(\mathcal{D}, \text{minsupp} = \frac{N\theta_1}{1-\theta_2})$ \\ $\triangleright$ mine candidate rules from $\mathcal{D}$ using minimum support from \textcolor{blue}{Theorem~\ref{theorem:support}}
            \FOR{$t=0\rightarrow T$}
            \STATE $\delta = \text{random}()$
            \IF{$\delta \leq \frac{1}{3}$ or $\Omega(\mathcal{R}_{[t]})\geq \frac{\lambda^*_{[t]} + \theta_2}{\theta_1}$} 
           \STATE $\mathcal{R}_{[t+1]} \leftarrow$ remove a rule from $\mathcal{R}_{[t]}$ $\triangleright$ decrease the size of $\mathcal{R}_{[t]}$ and use \textcolor{blue}{Theorem~\ref{theorem:dsize}}
            \ELSIF{$\delta \leq \frac{2}{3}$ or $\text{support}(\mR_{[t]})\leq \frac{\theta_1 - \lambda^*_{[t]}}{\theta_2}$} 
            \STATE $\mathcal{R}_{[t+1]} \leftarrow$ add a rule to $\mathcal{R}$ $\triangleright$ increase the transparency and use \textcolor{blue}{Theorem \ref{theorem:dexp}}
            \ELSE 
            \STATE $\epsilon \leftarrow \{k|f_{[t]}(\mathbf{x}_k) \neq y_k\}$ $\triangleright$ indices of misclassified examples.
            \IF{$\epsilon \in \text{covrg}(\mathcal{R}_{[t]})$}
            \IF{$\epsilon$ is negative}
            \STATE  $\mathcal{R}_{[t+1]} \leftarrow$ remove a rule from $\mathcal{R}_{+[t]}$ that covers $\epsilon$.
            \ELSE 
            \STATE $\mathcal{R}_{[t+1]} \leftarrow$ add a rule to $\mathcal{R}_{+[t]}$ to cover $\epsilon$ or remove a rule from $\mathcal{R}_{-[t]}$ that covers $\epsilon$ 
            \ENDIF
            \ELSE 
            \STATE $\mathcal{R}_{[t+1]} \leftarrow$ add a rule to $\mathcal{R}_{[t]}$ that is consistent with the sign of $\epsilon$
            \ENDIF
            \ENDIF 
            \STATE accept $\mathcal{R}_{[t+1]}$ with probability $\exp(\frac{\Lambda( \mathcal{R}_{[t]}) - \Lambda( \mathcal{R}_{[t+1]}) }{C_0^{\frac{t}{T}}})$ 
            \STATE $\mathcal{R}^* = \underset{\mathcal{R}_{[t+1]}, \mathcal{R}^*}{\arg\min} \Lambda(\mathcal{R})$ $\triangleright$ update the best solution
            \ENDFOR
    \STATE \textbf{Output:} $\mathcal{R}^*$
\end{algorithmic}
\end{algorithm}
\normalsize

\vspace{-2mm}
\paragraph{Rule Space Pruning} We first use FP-growth\footnote{FP-Growth is an off-the-shelf rule miner. Other rule miners such as Apriori or Eclat can also be used.}  to generate a set of candidate rules, $\Upsilon^+$ for positive rules and $\Upsilon^-$ for negative rules. The algorithm will search only within this rule space.  
 Since the number of rules grows exponentially with the number of features, the complexity of the algorithm is directly determined by the size of the rule space. To facilitate faster computation, we derive a lower bound on the support of rules to prune the rule space. All proofs are in the supplementary material.
\begin{theorem}[Lower Bound on Support]\label{theorem:support}
$\forall r \in \mR^*_+$, $\text{support}(r) \geq N\theta_1$; 
$\forall r \in \mR^*_-, \text{support}(r) \geq\frac{N\theta_1}{1-\theta_2}$.
\end{theorem}
This means $\mR^*$ does not contain rules with too small a support. This theorem is used before the search begins to prune the rule space to only contain rules with large support, excluding unqualified rules from consideration, which greatly reduces computation without hurting the optimality. On the other hand, removing rules with low support naturally helps prevent overfitting. The bound increases as $\theta_1$ increases, since $\theta_1$ represents the penalty for adding a rule.

\vspace{-2mm}
\noindent\textbf{Search Chain Bounding} We also derive two bounds that reduce the search space during the search. The bounds are applied in each iteration to confine the Markov Chain within promising solution space, preventing it from going too far and wasting too much search time. 
First, we derive a bound on the size of the model.
Let $\lambda^*_{[t]}$ represent the best objective value found till time $t$, i.e.
$$
\lambda^*_{[t]} = \min_{\tau \leq t}\Lambda(\mR_{[\tau]}).
$$\vspace{-2mm}
We claim
\begin{theorem}[Upper Bound on Size]\label{theorem:dsize}
$\Omega(\mathcal{R}^*) \leq \frac{\lambda^*_{[t]} + \theta_2}{\theta_1}$.    
\end{theorem}
This theorem says that the size of an optimal model is upper bounded, which means the Markov Chain only needs focus on solution space of small models. Therefore, in the proposing step, if the current state violates the bound, the next state should be proposed by removing a rule from the current model.

Next we derive an upper bound on the transparency with the similar purpose of confining the Markov Chain.
\vspace{-2mm}
\begin{theorem}[Lower Bound on Transparency]\label{theorem:dexp}
$\text{support}(\mR^*) \geq \frac{\theta_1 - \lambda^*_{[t]}}{\theta_2}$.    
\end{theorem}\vspace{-2mm}
The theorem says the transparency of an optimal model is lower bounded. Therefore whenever it is violated, the next proposal should be adding a rule to the current state.

Both bounds become smaller as $\lambda^*_{[t]}$ continuously gets smaller. Exploiting the theorem in the search algorithm, we check the intermediate solution at each iteration and pull the search chain back to promising an area (models of sizes smaller than $\frac{\lambda^*_{[t]} + \theta_1}{\theta_2}$ and models with transparency larger than $\frac{\theta_1 - \lambda^*_{[t]}}{\theta_1}$) whenever the bounds are violated (line 8 and line 10 in Algorithm 1). 

Now we detail the proposing step below.

\textbf{Proposing Step}: To propose a neighbor, at each iteration, we choose to improve one of the three terms (accuracy, interpretability, and transparency) with approximately equal probabilities. With probability $\frac{1}{3}$, or when the upper bound of the model in Theorem~\ref{theorem:dsize} is violated, we aim to decrease the size of $\mathcal{R}_{[t]}$ (improve interpretability) by removing a rule from $\mR_{[t]}$ (line 8 - 9). With probability $\frac{1}{3}$ or when the lower bound on transparency is violated, we aim to increase coverage of $\mR_{[t]}$ (improve transparency) by adding a rule to $\mR_{[t]}$ (line 10-11). Finally, with another probability $\frac{1}{3}$, we aim to decrease the classification error (improve accuracy) (line 13-22). 
To decrease the misclassification error, at each iteration, we draw an example from examples misclassified by the current model (line 13).  If the example is covered by $\mR_{[t]}$ (line 14), it means it was sent to the interpretable model but was covered by the wrong rule set. 
If the instance is negative, we remove a rule from the positive that covers it. If the instance is positive, we add a rule to $\mathcal{R}_{+[t]}$  to cover it or remove a rule in the $\mathcal{R}_{-[t]}$ that covers it, re-routing it to $\fb$. If the example is not covered by $\mR_{[t]}$, it means it was previously sent to the black-box model but misclassified, since we cannot alter the black-box model, we add a rule to the positive or negative rule set (consistent with the label of the instance) to cover the example, re-routing it to $\fin$.

When choosing a rule to add or remove, we first evaluate the rules using precision, which is the percentage of correctly classified examples of a rule. Then we balance between exploitation, choosing the best rule, and exploration, choosing a random rule, to avoid getting in a local minimum. 

\vspace{-2mm}
\section{Experiments}\label{sec:exp}
We perform a detailed experimental evaluation of HyRS using public datasets and a real-world application. We compare HyRS with state-of-the-art interpretable and black-box baselines. 
 
 \subsection{Experiments on Public Datasets}\label{sec:exp1}
The goal of the first set of experiments is to examine the accuracy and transparency of HyRS as well as their relationships. We use public datasets from domains interpretability is most pursed. 

\noindent\textbf{Datasets} We use four structured datasets and a text dataset from domains where interpretability is highly desired, including healthcare, judiciaries and customer analysis. 1) \emph{juvenile}\citep{osofsky1995effect} (4023 observations and 55 reduced features), to study the consequences of juvenile exposure to violence. The dataset was collected via a survey sent to juveniles.   2) \emph{credit card} (30k observations and 23 features), to predict the default of credit card payment \citep{yeh2009comparisons} 3) \emph{recidivism} (11,645 observations and 106 features) to predict if a criminal will re-offend after he is released from prison 4) \emph{readmission} (100,000 observations and 34 features) to predict readmission of patients with diabetes. 5) \emph{Yelp review} \citep{kotzias2015group} (1,000 observations) that contains positive and negative reviews from Yelp. The goal is to do sentiment classification.

\noindent\textbf{Implementation}
 We process each dataset by binarizing categorical features and discretizing real-valued features with four cut-off points. For each structured dataset, we build three black-box models that are often the top performing models, Random Forests \cite{liaw2002classification}, AdaBoost \cite{freund1995desicion} and extreme gradient boosting trees (XGBoost) \cite{chen2016xgboost}. For the text classification, we build a Long Short-Term Memory (LSTM) neural network that consists of one embedding layer with an embedding vector of 32, one layer of 100 LSTM units, and two fully connected layers following it.  
 We partition each dataset into 80\% training and 20\% testing. We do cross-validation for parameter tuning on the training set and evaluate the best model on the test set. The predictive performance of the black-box models are reported in Table \ref{tab:acc}.
 \vspace{-4mm}
 \begin{table}[h!]\renewcommand{\arraystretch}{1.1}
\centering
\caption{Test accuracy of black-box models }\label{tab:acc}
\small
\begin{tabular}{@{\hskip1pt}l@{\hskip1pt}@{\hskip1pt}c@{\hskip1pt}@{\hskip1pt}c@{\hskip1pt}@{\hskip1pt}c@{\hskip1pt}@{\hskip1pt}c@{\hskip1pt}@{\hskip1pt}c@{\hskip1pt}@{\hskip1pt}}
\toprule
Models    & \multicolumn{1}{@{\hskip3pt}c@{\hskip3pt}}{Juvenile }& \multicolumn{1}{@{\hskip3pt}c@{\hskip3pt}}{Credit card } & \multicolumn{1}{@{\hskip3pt}c@{\hskip3pt}}{Recidivism} & \multicolumn{1}{@{\hskip3pt}c@{\hskip3pt}}{Diabetes}& \multicolumn{1}{@{\hskip3pt}c@{\hskip3pt}}{Yelp} \\ \hline
 RF    & .91& .82  &.73& .64 &-- \\ 
AdaBoost & .90 & .82  &  .69   &.64&-- \\ 
XGBoost &  .90    &  .82     &.74& .64&-- \\ 
LSTM &--&--&--&--&0.76\\
\bottomrule
\end{tabular}
\end{table}\vspace{-2mm}
\normalsize

\subsubsection*{Study 1: Free Transparency}
The goal of the analysis is to find the maximally achievable ``\emph{free}'' transparency (no cost of the predictive performance), i.e., finding the area where a black-box is overkill.

We tune the parameters $\theta_1, \theta_2$ to get a set of models for each dataset. $\theta_1$ controls the number of rules and is chosen from [0.001, 0.01]. $\theta_2$ controls transparency and we choose $\theta_2$ from 0 to 1. We report in Table \ref{tab:transparency} the maximal free transparency. The performance is evaluated on test sets.
\vspace{-2mm}\begin{table}[h!]
\caption{Maximally achievable free transparency}\label{tab:transparency}
\footnotesize
\begin{tabular}{@{\hskip1pt}l@{\hskip1pt}@{\hskip1pt}c@{\hskip1pt}@{\hskip1pt}c@{\hskip1pt}@{\hskip1pt}c@{\hskip1pt}@{\hskip1pt}c@{\hskip1pt}@{\hskip1pt}c@{\hskip1pt}@{\hskip1pt}}
\toprule
 & Juvenile & Credit Card  & Recidivism & Diabetes & Yelp\\\hline
 $\langle \cdot, \text{RF}\rangle$& .82 & .91 &   .82 & .10 &--\\
 $\langle \cdot, \text{AdaBoost}\rangle$&.61 &.89  &  .79 & .26 &-- \\
  $\langle \cdot, \text{XGBoost}\rangle$&.78 &   .90  &  .65&  .20&--\\ 
   $\langle \cdot, \text{LSTM}\rangle$&--&--&--&--& .43\\ 
  \bottomrule      
\end{tabular}
\end{table}\vspace{-2mm}

The rules are able to explain on average roughly 80\% of the data for juvenile, credit card and recidivism, and about 20\% for diabetes dataset. The results show that our fundamental assumption is true - there exists a subspace where the interpretable model is as accurate than the black-box model, even if the black-box model is better globally. Meanwhile, this free interpretability is obtained using only a few rules (see Table \ref{tab:nrules}).  All models use less than 5 rules in total. For the Yelp dataset, we select the top 2000 words with highest frequency as the  input. 
\begin{table}[h!]
\caption{The number of rules in HyRS models}\label{tab:nrules}
\footnotesize
\begin{tabular}{@{\hskip2pt}l@{\hskip2pt}@{\hskip2pt}c@{\hskip1pt}@{\hskip1pt}c@{\hskip1pt}@{\hskip1pt}c@{\hskip1pt}@{\hskip1pt}c@{\hskip1pt}@{\hskip2pt}c@{\hskip2pt}@{\hskip2pt}}
\toprule
 & Juvenile & Credit Card  & Recidivism & Diabetes & Yelp\\\hline
 $\langle \cdot, \text{RF}\rangle$& 2 & 4&   2 & 4 &--\\
 $\langle \cdot, \text{AdaBoost}\rangle$&1 &3  &  2& 5 &-- \\
  $\langle \cdot, \text{XGBoost}\rangle$&1 &   4 &  1&  5&--\\ 
   $\langle \cdot, \text{LSTM}\rangle$&--&--&--&--&2\\ 
  \bottomrule      
\end{tabular}
\end{table}\vspace{-2mm}

\noindent\textbf{Examples of HyRS models} We show two examples of HyRS models built from the datasets above. The first example is from juvenile dataset. The data was collected from a survey so the features are questions and feature values are answers to that question. The positive rule set is an empty set, and the negative rule set consists of one rule.

\textbf{if} Has any of your family members or friends ever attacked you with a weapon $\neq$ Yes  \emph{and} Have your friends ever hit or threatened to hit someone without any reason? $\neq$ Yes \emph{and} Have your friends ever broken into a vehicle or building to steal something $\neq$ Yes \\
\textbf{then }$Y=0$\\
\textbf{else }$Y = \fb(\mathbf{x})$
 \begin{figure*}[h!]
\centering
  \includegraphics[width=0.85\textwidth]{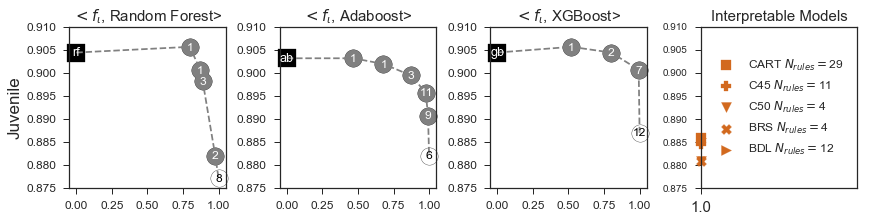}
\caption{The trade-off between transparency and accuracy for HyRS on Juvenile dataset. The black-squared represent black-box models. Grey circles represent HyRS models and the transparent circle represent HyRS models with transparency equal to one (reduced to interpretable models).}\label{fig:juvenile_ef}    \vspace{-2mm}
\end{figure*}

Note that this one rule captures 78\% of the instances and the overall predictive accuracy is just as accurate as of the black-box model. 

\begin{figure}[h!]
\centering
  \includegraphics[width=0.45\textwidth]{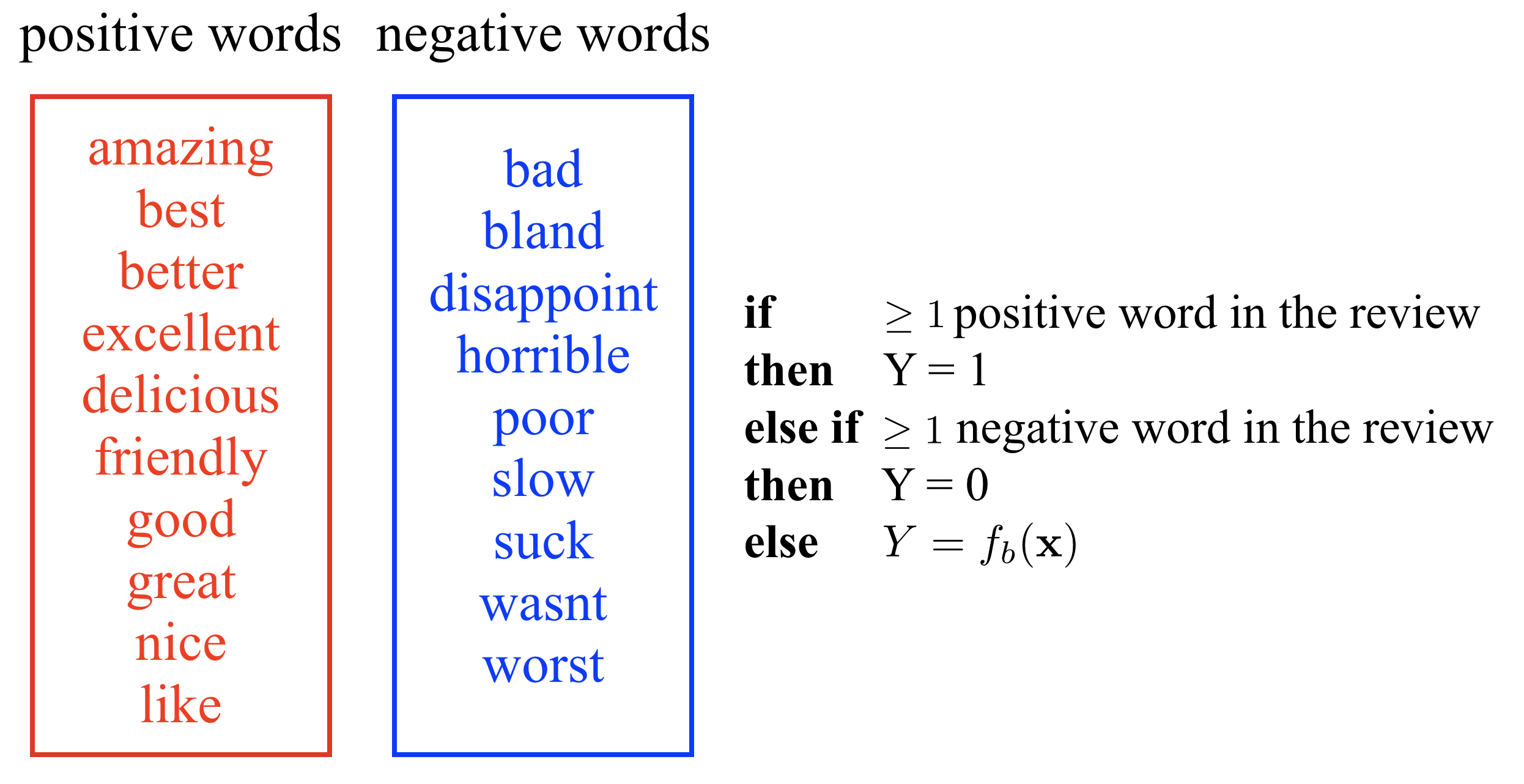}
\caption{Positive words and negative words mined from Yelp review dataset. The HyRS model built from these words captures 43\% of the reviews without losing predictive accuracy compared to the LSTM model.}\label{fig:words}\vspace{-6mm}   
\end{figure}

The second model we show is built from the text classification using the Yelp review data.  When we build HyRS on the text data, each unique word is a feature, and a rule is a phrase (conjunction of features), i.e., word 1 and word 2 and $\cdots$ all appear in the text. Therefore, a rule set contains a set of words and phrases. On this dataset, the rules are all words. We show the HyRS model and the words for the positive set and the negative set in Figure \ref{fig:words}.

\subsubsection*{Study 2: Accuracy-Transparency Trade-off}
In addition to studying the maximally achievable ``free''  transparency, we study if some loss in predictive performance is tolerable, how can that be efficiently utilized and transformed into transparency. To visualize this trade-off, we plot models' transparency as the X-axis and accuracy as the Y-axis, both evaluated on test sets. Here we choose juvenile dataset for demonstration. We tune $\theta_1$ from 0.001 to 0.01 and $\theta_2$ from 0 to 1 and only show the models on the frontier of the curves. See Figure \ref{fig:juvenile_ef}. 

\noindent\textbf{Baselines} We benchmark the performance of HyRS against other rule-based interpretable models, C4.5 \cite{quinlan2014c4} and C5.0 \cite{kuhn2014c50}, Scalable Bayesian Rule Lists (SBRL) \cite{ynormalize_addang2016scalable} and Bayesian Rule Sets (BRS) \cite{wang2017bayesian}. BRS and SBRL are two recent representative methods which have proved to achieve simpler models with competitive predictive accuracy compared to the older rule-based classifiers.   See a description of parameter tuning in the supplementary material. We find the models with the highest cross-validated accuracy and show their test accuracy in Figure \ref{fig:juvenile_ef}. All models are at the transparency of 1 since they are interpretable.

The curves remain almost flat when the transparency is smaller than roughly 70\% and start to drop quickly after that. It means there exists a large subset of data that can be captured by simple rules (only one or two as shown in Figure \ref{fig:juvenile_ef}). The rules perform as well as black-box models. But the remaining 30\% is much harder to be characterized by rules. Thus the number of rules increases quickly as transparency approaches 1, and the accuracy drops to values comparable to the interpretable baselines. 
  \begin{table*}[h!]
\centering
\small
\caption{A HyRS model for predicting  outcomes for medical crowdfunding. The accuracy is 0.77 and transparency is 0.82.}
\label{tab:med}
\begin{tabular}{lll}
\toprule
        & \multicolumn{1}{c}{\textbf{Rules}}                                                             & \multicolumn{1}{c}{\textbf{Model}}             \\\hline
\textbf{if}      & day of launch (of a month) $< 6$ \emph{and} approval  status $\neq$ approved  \emph{and} fundraiser's gender = missing  &  \\
        & OR month of launch = June& \\
        & OR day of launch (of a month) $< 18$ & \\
        & OR content length $<$ 519 words \emph{and} target amount $\geq$ 44,000 &\\
\textbf{then} & \textcolor{red}{$Y=1$ (failure) }            &    $\mathcal{R}_+$                                   \\
\textbf{else if} & content length $<$ 334 words \emph{and} patient age $<$ 44 \emph{and} target amount $<$44,000 &             \\
& OR approval  status = approved \emph{and} target amount $\leq$ 29,000 \emph{and} month $\neq$ June &               \\
& OR content length $\geq$ 519 words \emph{and} patient age $<$ 44 \emph{and} month $\neq$ June &               \\
\textbf{then} & \textcolor{red}{ $Y = 0$ (success)} &    $\mathcal{R}_-$                                   \\
\textbf{else}    & \textcolor{blue}{$Y = \fb(\mathbf{x})$  }                          &  $f_b$      \\ \bottomrule                
\end{tabular}
\end{table*}
\subsection{Application to Medical Crowdfunding Prediction}
We apply HyRS to a real-world application, medical crowdfunding prediction, using real-world data. Medical crowdfunding is a type of donation-based crowdfunding, helping users raise funds to pay medical bills by collecting small donations from many people. A crowdfunding could last for several weeks.  But since medical crowdfunding is time-critical in its nature, an early prediction of its outcome, especially future failure is very valuable since users can look for alternatives channels of funding early if they could not raise enough money from the current platform. We want to predict the failure of fundraising, which is defined as raising less than 10\% of the target amount. The data is provided by medical crowdfunding company and consists of 51,228 cases from October 2016 to June 2018.

There are two types of features, time-invariant and time series. When a fundraiser creates a new case, he submits patient information including demographics (age, gender, city, etc.), insurance status (commercial or basic medical insurance), the target amount,  verification from the patient's hospital,  along with a short ``call for donation'' post.  These are time-invariant features. Then, after a case is published, a set of features are collected daily. The daily features include the number of times a case is shared on social networks, the number of times a fundraiser responds to users' questions and requests for more information (or proof) on the case pages, the number of views for the case, the number of users who verified the case, the number of users who donate, and etc. These are time series features that are collected daily. We use the observations for the first week after a case is published.

We design a deep neural network that first uses two layers of LSTM (20 units and 40 units) to process the time series features, and then the output is merged with the time-invariant features at two fully connected layers. The last layer uses a sigmoid activation function to produce a  prediction for failure. We partition the dataset into 80\% training and 20\% testing. We train the network for 200 epochs, and the test accuracy is 0.96. 

Then we build a HyRS model. To ensure interpretable and early prediction, here we restrict our model only to use time-invariant features that are immediately available after a user launches a case. We would like to know how accurately can HyRS provide an early prediction only using a few features.

\begin{figure}[h!]
\centering
  \includegraphics[width=0.36\textwidth]{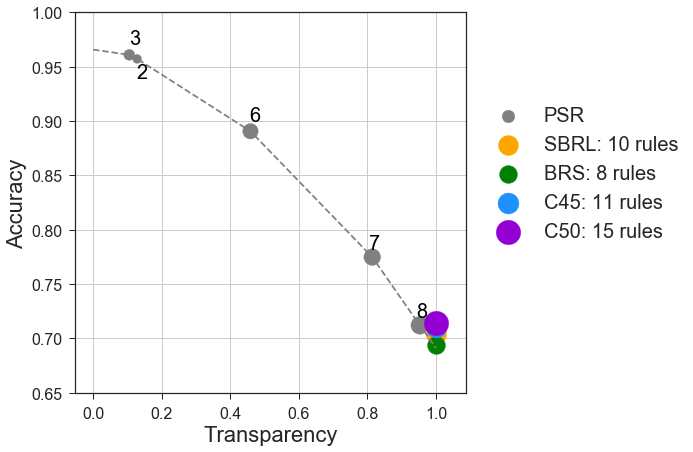}
\caption{The accuracy vs transparency curve for HyRS models for predicting fundraising failure for  medical crowdfunding.}\label{fig:tradeoff}    
\end{figure}

To understand the trade-off between transparency and accuracy, we set $\theta_1$ to 0.001 and vary $\theta_2$ from 0 to 1 to obtain a curve in Figure \ref{fig:tradeoff}. The number of rules is represented by the size of the markers and also annotated in the figure.

\vspace{-2mm}

 For the baseline models, we use the same interpretable models as described in Section \ref{sec:exp1}. We follow the same steps tuning the parameters. Their accuracy and the number of rules are shown in Figure \ref{fig:tradeoff}.
 
 \noindent\textbf{Discussions}  This curve characterizes the trade-off between transparency and accuracy. As transparency increases, HyRS covers more and more data with rules, at the cost of predictive accuracy, but always higher than pure interpretable models alone. The users can decide the best operating point based on their specific requirement of accuracy using the curve. Compared to using either purely black-box models or purely interpretable models, HyRS provides more options for model selections.
 
  We show a model that achieves transparency of 0.82 and an accuracy of 0.77 in Table \ref{tab:med}. This model consists of four positive rules and three negative rules. Only 18\% instances are not captured and therefore sent to LSTM for decision.

\vspace{-2mm}
\section{Conclusions}\label{sec:conclusion}
We proposed a general framework for learning a hybrid model that integrates an interpretable partial substitute with \emph{any} black-box model to introduce transparency into the predictive process at no or low cost of the predictive performance. We instantiated this framework with Hybrid Rule Sets Hybrid (HyRS) model using rules as the interpretable component. 
Experiments demonstrated partial transparency is possible in the presence of black-box models. It suggests that in some cases, always using a black-box is overkill, and replacing with a simpler and interpretable model will save resources and provide partial transparency.

The HyRS model is one example of the proposed hybrid model framework. An important contribution of this work is that we proposed a general framework for combining an interpretable substitute model with a black-box model. Our framework can support the exploration of a variety of interpretable models, such as linear models (see our latest paper \cite{wanghybrid2019}), decision trees and prototype-based models.

The proposed framework provides a new solution when one wishes not to give up the high predictive accuracy of black-box models. A hybrid model can serve as a pre-step for a black-box explainer: we first find a region that can be captured and explained by an interpretable model and then sends the rest of the data to a black-box to predict and an explainer to explain.

\textbf{Code} for HyRS is available at \\ \url{https://github.com/wangtongada/HyRS}

\bibliography{tong,iml,nips_msr,jmlr_msr,rules_ICDM}
\bibliographystyle{icml2019}
\end{document}